\documentclass[a4paper]{article}
\usepackage{jsaisig}
\usepackage{subcaption}
\usepackage{url}
\usepackage{amsmath}
\usepackage{algorithmic}
\usepackage{algorithm}
\usepackage{cite}
\usepackage{subcaption}
\usepackage{fancyvrb}

\usepackage[pdftex]{graphicx}

\usepackage{appendix}
\usepackage{amssymb}

\newcommand{\bn}{\textsf{BeliefNest}}

\begin{document}
\title{BeliefNest: A Joint Action Simulator \\ for Embodied Agents with Theory of Mind}

\author{Rikunari Sagara\afil{1} \quad Koichiro Terao\afil{2} \quad Naoto Iwahashi\afil{2}}
\affiliation{%
 \afil{1}University of Shizuoka \quad \afil{2}Okayama Prefectural University}

\abstract{
This paper introduces an open-source simulator, \bn, designed to enable embodied agents to perform collaborative tasks by leveraging Theory of Mind. \bn\ dynamically and hierarchically constructs simulators within a Minecraft environment, allowing agents to explicitly represent nested belief states about themselves and others. This enables agent control in open-domain tasks that require Theory of Mind reasoning. The simulator provides a prompt generation mechanism based on each belief state, facilitating the design and evaluation of methods for agent control utilizing large language models (LLMs). We demonstrate through experiments that agents can infer others' beliefs and predict their belief-based actions in false-belief tasks.
}

\maketitle
\thispagestyle{empty}

\begin{figure}[htbp]
\begin{center}
\includegraphics[width=140mm]{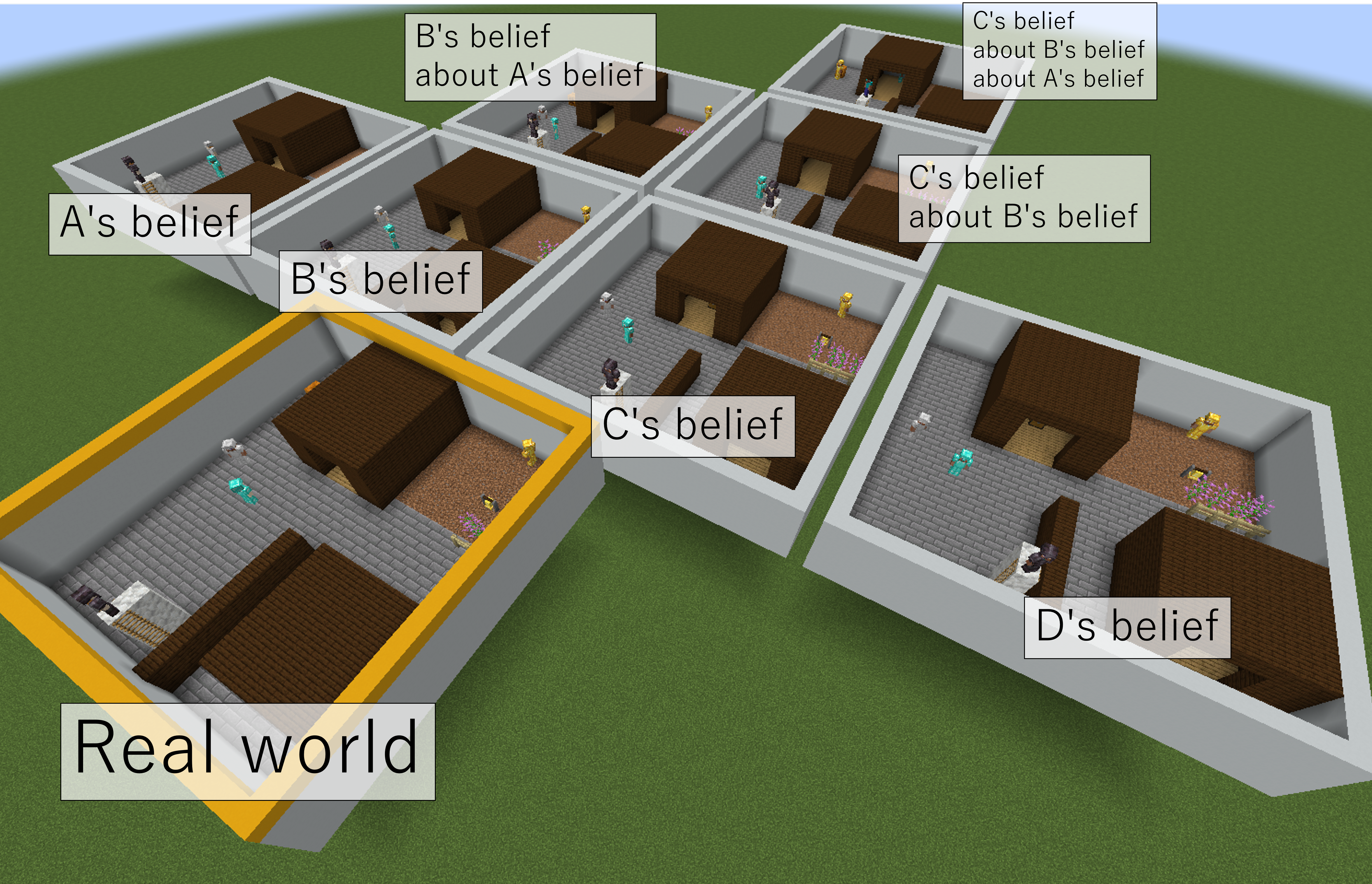}
\caption{Overview of \bn. The nested belief structure derived from Theory of Mind is implemented using multiple physical simulators. Each agent can perform mental simulations within its own simulator, considering the beliefs of others, to generate action plans that reflect these beliefs.}
\label{fig:overview}
\end{center}
\end{figure}

\section{Introduction}
Theory of Mind is a fundamental cognitive ability that underpins human social behavior, enabling individuals to infer the beliefs, intentions, and knowledge of others. In this paper, we propose \bn, an open-source simulator designed to support research on collaborative behavior in embodied agents endowed with Theory of Mind capabilities. Recent advances in embodied agents powered by large language models (LLMs) have shown promising progress. However, there is still no platform that can explicitly represent nested belief states and integrate them with action generation mechanisms. \bn\ addresses this gap by providing a flexible simulation framework that incorporates both hierarchical belief structures and prompt generation support.

\bn\ offers the following features:
\begin{itemize}
\item Explicit representation of nested belief states, as studied in Theory of Mind, using hierarchical simulators (see Section~\ref{section:nest})
\item Support for prompt generation based on each belief state, enabling the design and evaluation of methods for agent control with LLMs (see Section~\ref{section:control})
\item Integration with the Minecraft environment, which is widely used in LLM agent research~\cite{wang2023voyager, yocum2023-multivoyager, altera2024sid, cai2025rocket2}, and support for open-domain tasks
\end{itemize}

In this paper, we describe the design and functionality of \bn\ and demonstrate its effectiveness through experiments on false-belief tasks. The source code is available at \url{https://github.com/sagara-r/BeliefNest}.

\section{Related Work}

Theoretical frameworks involving hierarchical belief structures, such as ``X thinks that Y thinks that ...'', have been widely discussed in dialogue and planning research that leverages Theory of Mind \cite{Taylor1996-zq, ho2022planning}. In addition, numerous studies have aimed to explicitly model such nested beliefs to enhance the efficiency of dialogue and planning systems \cite{McEleney2005-uz, Muise2021-sv, Cohen2023-oz}. More recently, growing attention has been given to evaluating the extent to which LLMs can perform Theory of Mind reasoning \cite{sap2022-cc, ullman2023large, xu2024opentom, street2024-se, kosinski2024tom, strachan2024tom}, as well as to developing methods for improving ToM capabilities in LLMs \cite{Qiu2023-ib, zhang2023building, oguntola2023-rm, wilf2024thinktwice, ying2024goma, zhang2025autotom}. Further, several works~\cite{wilf2024thinktwice, Ma2025-gr, yeh2025perspective} have addressed tasks that require distinguishing between different perspectives, an aspect central to Theory of Mind. In particular, Yeh et al.~\cite{yeh2025perspective} highlight the limitations of LLMs in distinguishing belief states grounded in differing perspectives. The \bn\ simulator proposed in this study aims to support Theory of Mind-based reasoning by systematically organizing belief state information to be fed into LLMs, thereby addressing these limitations.

\section{Nested Structure of the Simulator}
\label{section:nest}

\begin{figure}[!htbp]
\begin{center}
\includegraphics[width=90mm]{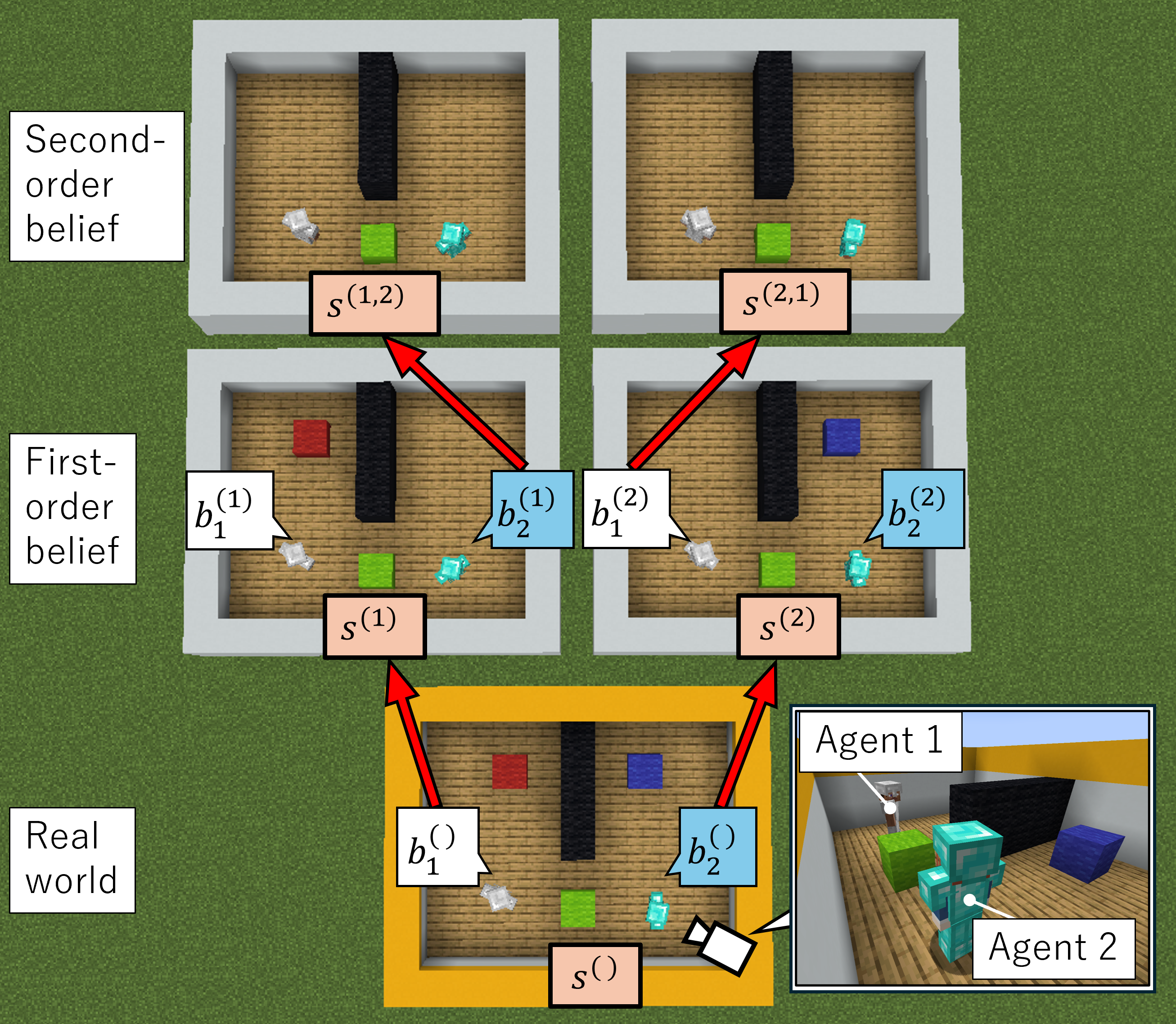}
\caption{Overall structure of \bn. The simulators form a nested hierarchy that corresponds to the nested structure of beliefs. Let $Z$ be a finite sequence of agent identifiers that represents a path of belief nesting. In the simulator representing state $s^{Z}$, agent $i$ constructs a new simulator whose state $s^{Z||i}$ reflects its belief $b^Z_i$. Here, the operator $||$ denotes sequence concatenation: $Z||i$ is the sequence obtained by appending agent $i$ to the end of $Z$. In the real world, agent 1 can see the red and green blocks, while agent 2 can see the blue and green blocks. Agent 1 is unaware of the blue block, and agent 2 is unaware of the red block. Therefore, agent 1 believes that agent 2 can see only the green block. These belief-dependent perceptions are explicitly represented within each simulator. All agents are assumed to have prior knowledge of the floor, the white outer walls, and the black inner wall.}
\label{fig:nest}
\end{center}
\end{figure}

Figure~\ref{fig:nest} illustrates the overall architecture of \bn. The system consists of a ``real-world simulator,'' which represents the actual environment, and multiple ``belief simulators'' that hierarchically represent the belief structures of agents. All of these simulators are instantiated within the Minecraft world, each occupying a distinct and isolated region of the environment. Each agent possesses a dedicated belief simulator that represents its own beliefs and can dynamically construct additional layers as needed. This allows for the explicit representation of nested belief states such as ``myself in the mind of another'' or ``someone else in the mind of another.'' Belief simulators can be dynamically added or removed depending on the context. In this paper, we refer to simulators in which an agent is physically present as shallower-level simulators, and to those that reflect the beliefs held by that agent as deeper-level simulators.

\section{Observation and Belief Updating}

Each belief simulator reflects the belief state based on observations made by agents in shallower-level simulators. However, when an agent engages in planning or hypothesis testing, it must temporarily suspend belief updates based on shallower-level simulators and instead generate its own timeline within the belief simulator to perform inference. Here, a ``timeline'' refers to a sequence of observations and actions recorded over time. The ability to generate such internal timelines corresponds to mental simulation, which is essential for planning. To support this functionality, we define two operational modes for simulators:
\begin{enumerate}
\item \textbf{Control mode}: The agent performs actions and generates a virtual timeline.
\item \textbf{Follow mode}: The simulator passively updates its state by sequentially reflecting belief information from the shallower-level simulator.
\end{enumerate}

Each simulator can switch modes independently. The pseudocode of the asynchronous belief updating process for each simulator is shown in Algorithm~\ref{alg:simulator}. Below, we describe the main processes performed in each mode.

In control mode, agents act within the simulator, and environment and agent information at each time step is recorded. The simulator performs the following C- processes, which are specific to control mode:

\begin{description}
\item[C-1: State acquisition] The current state $s$, representing the state of the environment and the agents, is obtained as objective information. Since state transitions are managed internally by the Minecraft environment, \bn\ does not track them directly; thus, this step retrieves the current state as needed.
\end{description}

In follow mode, the simulator receives and reflects belief states sent from the shallower-level simulator. The simulator performs the following F- processes, which are specific to follow mode:

\begin{description}
\item[F-1: Receiving belief state from shallower-level simulator] Belief information sent via step (Common-3) in the shallower-level simulator is received and adopted as the current state.
\item[F-2: Reflecting state in the simulator environment] The received state is used to update the environment and agent configurations within the simulator.
\end{description}

In addition, as common processing, the following steps are performed for each agent $i \in \mathcal{A}$, where $\mathcal{A}$ denotes the set of agents in the simulator.
\begin{description}
\item[Common-1: Observation based on perspective] Subjective observations $o_i$ are obtained via the observation function $O_i$, which captures visible elements in the environment and other agents from the agent's viewpoint.
\item[Common-2: Belief updating] The belief state $b_i$ is updated using observation $o_i$. Environmental and agent-related information is updated for observable elements, while unobserved elements remain unchanged.
\item[Common-3: Propagation to deeper-level simulators] The updated belief state is sent to the belief simulators associated with each agent.
\end{description}

As an example, when the real-world simulator is in control mode and all belief simulators are in follow mode, agent actions in the real-world simulator propagate belief updates to each belief simulator, and the resulting changes are reflected in their internal environments. Subsequently, if a particular belief simulator switches to control mode, belief updates from the real world are no longer applied to it. The agent's actions within that simulator become a hypothetical simulation, exerting no influence on the real world.

\begin{algorithm}[H]
	\caption{Asynchronous belief update process on each simulator}
	\label{alg:simulator}
	\begin{algorithmic}[1]
    \STATE \textbf{Initialization:} $b_{i} \quad \forall i \in \mathcal{A}$ 
    \WHILE{true}
        \IF{$\text{mode} = \text{``control''}$}
            \STATE $s \leftarrow \text{get\_state()}$ \dotfill \textbf{(C-1)}
        \ELSIF{$\text{mode} = \text{``follow''}$}
            \STATE $s \leftarrow \text{await receive\_belief()}$ \dotfill \textbf{(F-1)}
            \STATE $\text{apply\_to\_simulator}(s)$ \dotfill \textbf{(F-2)}
        \ENDIF
        \FORALL{$i \in  \mathcal{A}$}
            \STATE $o_{i} \leftarrow O_{i}(s)$ \dotfill \textbf{(Common-1)}
            \STATE $b_{i} \leftarrow \text{update\_belief}(b_{i}, o_{i})$ \dotfill \textbf{(Common-2)}
            \STATE $\text{send\_belief}(i, b_{i})$ \dotfill \textbf{(Common-3)}
        \ENDFOR
    \ENDWHILE

	\end{algorithmic}
\end{algorithm}

\section{Agent Control Based on Belief}
\label{section:control}

Agents within any simulator can be controlled via JavaScript programs. Since the state of each belief simulator reflects the beliefs of the agents in the corresponding shallower-level simulator, it is possible to generate JavaScript programs using LLMs based on the simulator's state, enabling agents to take actions grounded in their beliefs. To support program generation by LLMs, \bn\ provides functionality for prompt generation based on the state of each simulator. When a prompt template written in Jinja2 format (see Appendix~\ref{appendix:template} for an example) is provided, \bn\ automatically inserts a string representation of the simulator's state into the specified locations within the template to generate the final prompt. Users are free to employ any prompt template or LLM of their choice, making it possible to design and evaluate various methods for agent control utilizing LLMs.

As control primitives, a set of functions is provided to enable agent actions. These include, for example, movement, block breaking, item crafting, and taking from or storing items into chests. These capabilities are implemented using the Mineflayer library. We build upon and modify the action primitives introduced in Voyager \cite{wang2023voyager} to better fit our objectives.

\section{Timeline Management}
\label{section:branch}

Each simulator implements a mechanism for managing timeline branches, which represent divergent temporal progressions. Each branch can be assigned a unique identifier, and users can switch to any existing branch at any time. When a branch is selected, the simulator reverts to the latest state recorded in that branch, allowing the timeline to proceed from that point onward. If it is necessary to return to a specific point in the past, a branch can be created at that time in advance, enabling later restoration to that state.

By leveraging the branching mechanism, users can freely design and evaluate control strategies in which agents hypothetically explore multiple action plans based on others' beliefs before selecting one for actual execution. The specific procedures, such as how trials are conducted or how prompts are structured, are left to the user, while \bn\ provides a flexible simulation framework to support such exploratory processes.

\section{Example 1: Sally-Anne Task}

The Sally-Anne task \cite{Baron-Cohen1985sally} is a well-known example of a first-order false-belief task, widely used to assess whether an individual can infer beliefs that differ from their own. In this experiment, we evaluate whether a particular agent (hereafter referred to as the observer) can infer the beliefs and belief-based actions of other agents. Notably, the observer is not an external third party but an in-world agent embedded within the environment. Through the belief simulator constructed by \bn, the observer is expected to correctly infer Sally's belief and the actions derived from it.

\begin{figure}[h]
  \centering
  \begin{subfigure}[t]{0.3\columnwidth}
    \centering
    \includegraphics[width=\columnwidth]{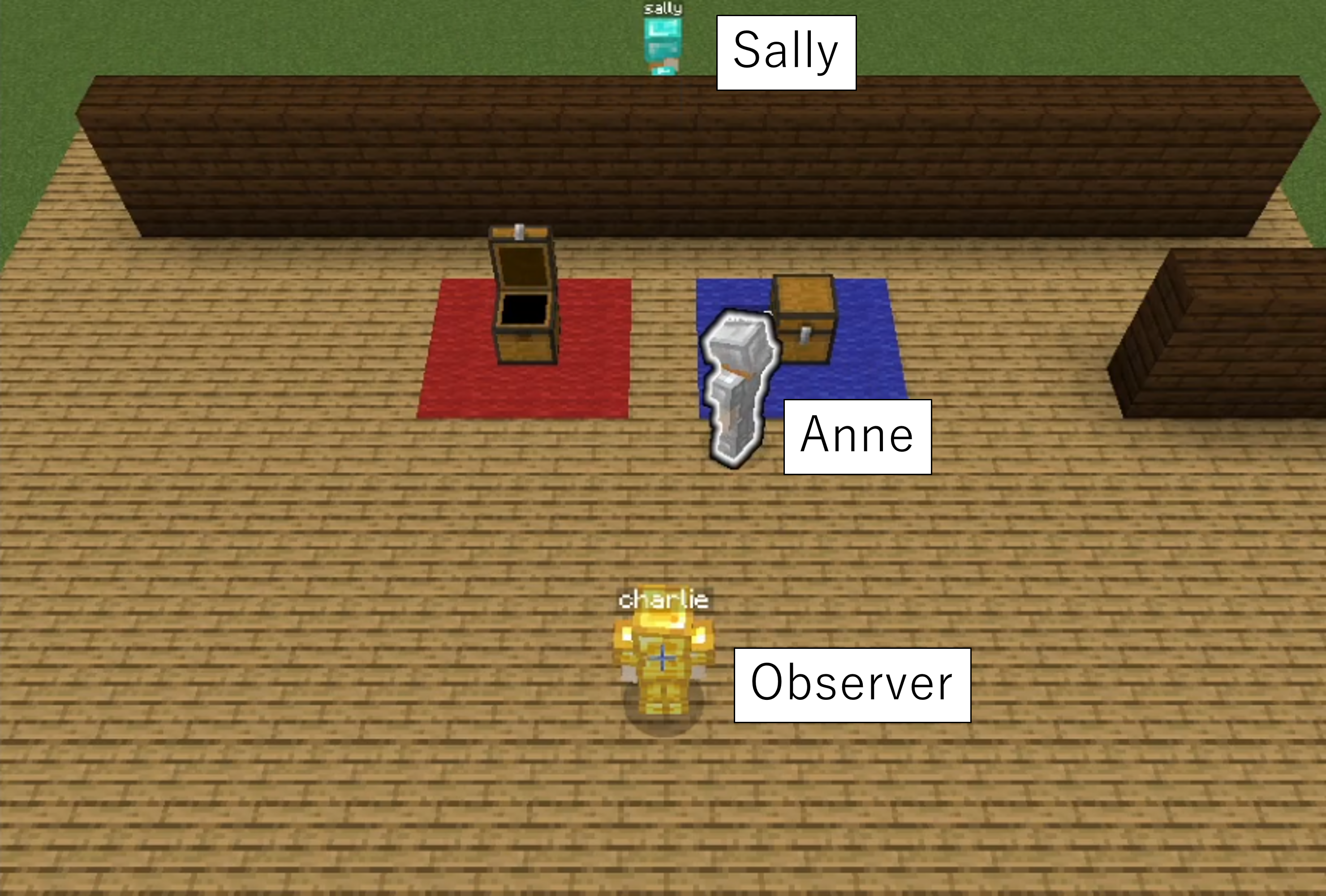}
    \caption{Condition 1.}
    \label{fig:sally1}
  \end{subfigure}
  \hspace{5mm}
  \begin{subfigure}[t]{0.3\columnwidth}
    \centering
    \includegraphics[width=\columnwidth]{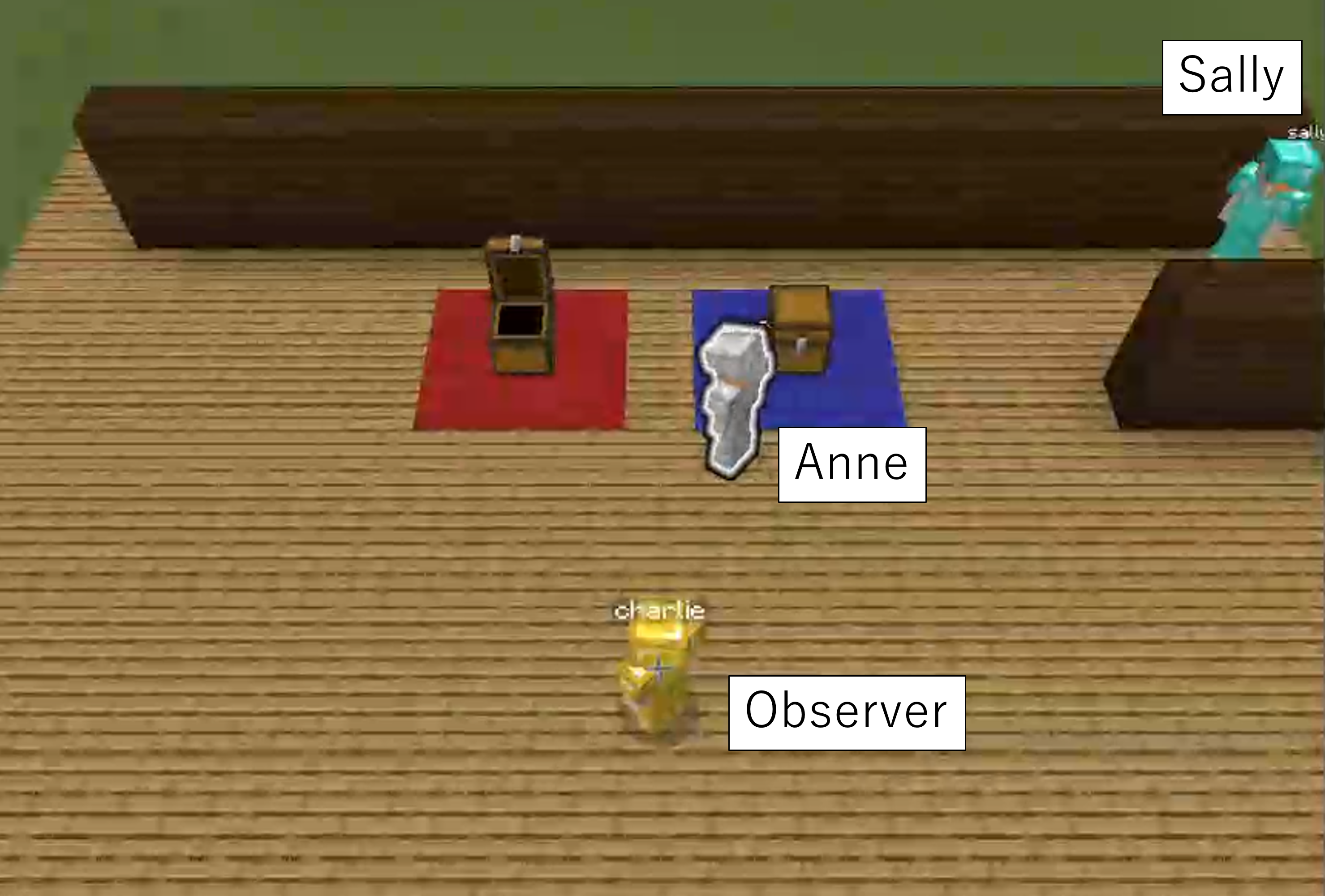}
    \caption{Condition 2. The observer can see Sally.}
    \label{fig:sally2}
  \end{subfigure}
  \hspace{5mm}
  \begin{subfigure}[t]{0.3\columnwidth}
    \centering
    \includegraphics[width=\columnwidth]{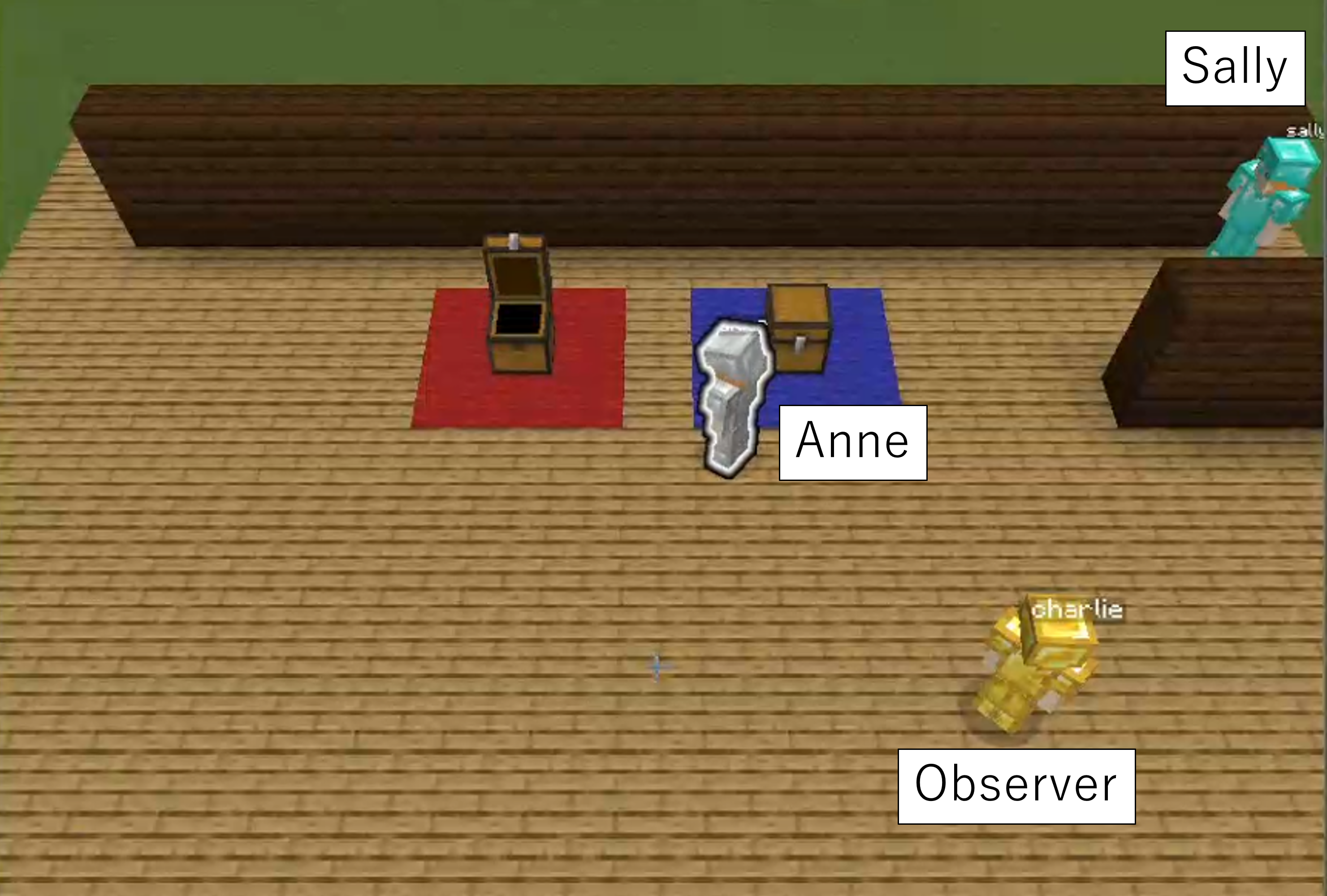}
    \caption{Condition 3. The observer cannot see Sally.}
    \label{fig:sally3}
  \end{subfigure}
  \caption{Experimental setup for the Sally-Anne task. Sally's position in the figure represents her location after placing the diamond in the box.}
  \label{fig:sally_all}
\end{figure}

Figure~\ref{fig:sally_all} shows the experimental setup, which includes three players (Sally, Anne, and the observer) and two boxes. In this task, subtle differences in the spatial configuration of agents can lead to varying visibility conditions and thus to multiple plausible interpretations. We define the following three conditions to capture these variations:

\begin{itemize}
\item \textbf{Condition 1 (standard setup)}: Sally places a diamond into the left box and then moves behind a wall. Anne subsequently transfers the diamond to the right box, and Sally does not observe this action. The observer witnesses the entire sequence.
\item \textbf{Condition 2}: Unlike Condition 1, Sally moves to a location where she can observe Anne's action. The observer sees the entire sequence as well.
\item \textbf{Condition 3}: As in Condition 2, Sally moves to a position where she can see Anne's action, but the observer cannot see where Sally moved, and therefore does not know that Sally observed the action.
\end{itemize}

A first-order false-belief task measures whether an agent can infer another agent's first-order belief. In this setting, the observer constructs a belief about Sally's belief, which corresponds to a second-order belief from the observer's perspective. This experiment evaluates whether the second-order belief simulator is constructed correctly, thereby enabling the observer to infer Sally's belief. Furthermore, based on the constructed belief state, the Sally agent within the belief simulator is given the task ``Get a diamond from a chest.'' We then evaluate whether the generated action is consistent with the inferred belief state, using GPT-4o \cite{OpenAI2024-ss} to generate actions based on prompts informed by belief states.

In the simulator representing the observer's first-order belief (i.e., the observer's own belief), the diamond was moved to the right box in all conditions. In contrast, the second-order belief simulator (i.e., the observer's belief about Sally's belief) produced different results depending on the condition. The inferred beliefs and actions were as follows:

\begin{itemize}
\item \textbf{Condition 1}: Reflecting the fact that Sally did not observe Anne's action, it was inferred that she believed the diamond was still in the left box. Consequently, it was predicted that Sally would open the left box.

\item \textbf{Condition 2}: Since the observer recognized that Sally had observed Anne's action, the second-order belief also inferred that Sally believed the diamond was in the right box. Therefore, it was predicted that Sally would open the right box.

\item \textbf{Condition 3}: Although Sally actually saw Anne's action, the observer could not confirm this. As a result, in the second-order belief, it was inferred that Sally believed the diamond remained in the left box, and thus would open the left box.
\end{itemize}

These results demonstrate that \bn\ is capable of accurately constructing others' beliefs that differ from an agent's own, based on subtle differences in spatial configuration. Moreover, it enables appropriate inference of belief-based actions in accordance with those constructed beliefs.

\section{Example 2: Ice Cream Van Task}

The Ice Cream Van Task \cite{Perner1985icecream} is a second-order false-belief task that involves a two-level belief hierarchy (e.g., X believes that Y believes that Z), and is used to evaluate whether an agent can correctly handle deeper belief structures. In this experiment, we examine whether the observer can infer ``what John thinks Mary believes.''

\begin{figure}[htbp]
\begin{center}
\includegraphics[width=70mm]{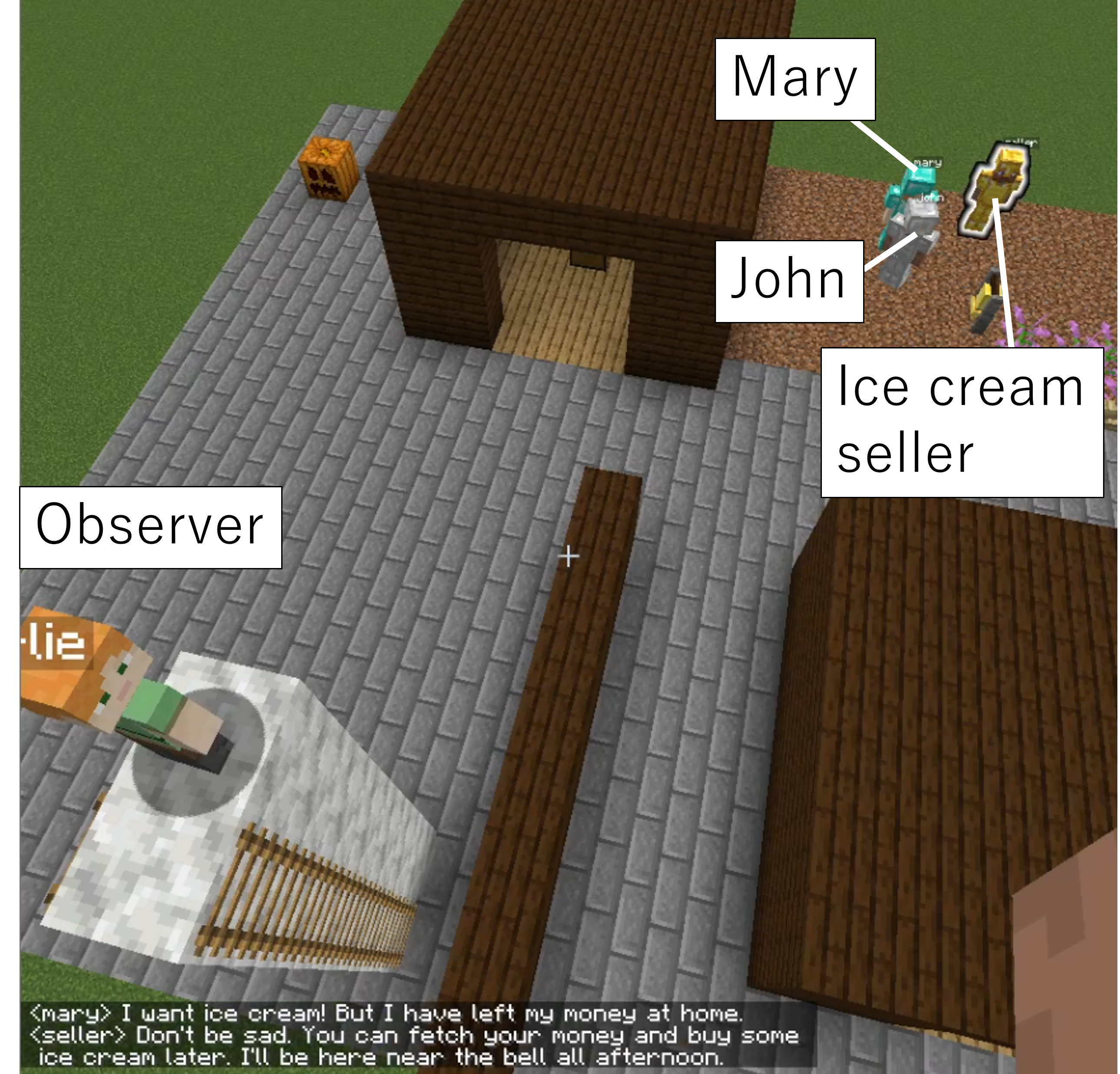}
\caption{Experimental setup for the Ice Cream Van Task.}
\label{fig:icecream}
\end{center}
\end{figure}

Figure~\ref{fig:icecream} illustrates the experimental setup, which includes four players: John, Mary, the ice cream seller, and the observer. The scenario unfolds as follows:
\begin{enumerate}
\item The ice cream seller tells both John and Mary at location A that he will stay there today.
\item Mary returns home.
\item The ice cream seller then tells John that he is going to location B, and leaves location A.
\item \label{itm:4} On the way, the ice cream seller unexpectedly encounters Mary and tells her as well that he is going to location B.
\item John returns home.
\end{enumerate}
The observer witnesses the entire sequence. The goal is to determine whether the observer can accurately infer John's second-order belief by verifying whether the third-order belief simulator (i.e., observer's belief about John's belief about Mary's belief) is correctly constructed.

Based on the resulting belief state, the Mary agent within the belief simulator is given the task ``Go to the ice cream seller. The experiment then evaluates whether the generated action is appropriate with respect to the inferred belief.

In the observer's third-order belief simulator, Mary was inferred to go to location B. This inference is based on the observer's recognition that John was unaware of Mary's reunion with the ice cream seller in Step~\ref{itm:4}. This result aligns with the expected reasoning for this task, demonstrating that \bn\ can accurately represent complex belief structures and generate plausible belief-based behavior.

\section{Conclusion}

In this paper, we introduced \bn, an open-source simulator designed to support research on collaborative behavior by equipping embodied agents with Theory of Mind capabilities. Through experiments, we demonstrated that agents using \bn\ can correctly infer others' beliefs and predict their actions in false-belief tasks. While this study focused on false-belief scenarios to validate the core functionalities of \bn, the simulator is applicable to a broader range of research on joint action. We expect that \bn\ will serve as a valuable foundation for advancing studies on collaboration in dialogue systems and robotics.



\bibliographystyle{unsrt}
\bibliography{bibliography}

\newpage
\section*{Appendix}
\appendix

\section{Prompt Generation Support by \bn}

\subsection{Example of a Prompt Template}
\label{appendix:template}

As an example, when the following template is provided to \bn, placeholders such as \verb#$$LAST_CODE$$# and \verb#{{ branch | chatlog }}# are replaced with string representations of the simulator state. Elements like \verb#chat_log# are Jinja2 filters, which users may also customize or extend as needed. The variable \verb#branch# stores a string that identifies the target simulator and branch (see Section~\ref{section:branch}). It is also possible to include information from multiple simulators and branches within a single prompt.

\vspace{3mm}

\begin{Verbatim}[frame=single]
Code from the last round:
$$LAST_CODE$$

Error from the last round:
$$LAST_ERROR$$

<Information>
Thought:
{{ branch | thought }}

Chat log:
{{ branch | chat_log }}

Position:
{{ branch | position }}

Chest:
{{ branch | chests }}

Inventory:
{{ branch | inventory }}

Other players:
{{ branch | other_players }}

Blocks seen so far:
{{ branch | blocks(["chest"]) }}

<History of player visibilities and actions>
[My(sally's) perspective]
{{ branch | events_and_visibilities }}

Task: {{ task }}
\end{Verbatim}

\subsection{Example of a Generated Prompt}
\begin{Verbatim}[frame=single]
Code from the last round:
No code was executed

Error from the last round:
No error

<Information>
Thought:
No thought

Chat log:
No chats

Position:
[-4.551, -51, -5.871]

Chest:
(-2, -51, -4): {'diamond': 1}
(2, -51, -4): {}

Inventory:
No data

Other players:
{
  "anne": {
    "position": "Cannot be seen",
    "helditem": "iron_chestplate",
    "inventory": "No data"
  }
}

Blocks seen so far:
chest visibilities:{
  "(-2, -51, -4)": {
    "Me": {
      "seen_before": true,
      "visible_now": false
    }
  },
  "(2, -51, -4)": {
    "Me": {
      "seen_before": true,
      "visible_now": false
    }
  }
}
lever visibilities: Not observed

<History of player visibilities and actions>
[My(sally's) perspective]
time;action;agent_name;description
69;depositItemIntoChest;sally;chest:(-2, -51, -4) & items:{'diamond': 1}

Task: Get a diamond from a chest.
\end{Verbatim}

\end{document}